\documentclass[conference]{IEEEtran}
\IEEEoverridecommandlockouts

\usepackage{mathtools}
\usepackage{multirow}

\usepackage{float}
\usepackage{multirow}
\usepackage{amsmath}
\usepackage{lipsum}
\usepackage{svg}

\usepackage{cite}
\usepackage{amsmath,amssymb,amsfonts}
\usepackage{algorithmic}
\usepackage{graphicx}
\usepackage{textcomp}
\usepackage{xcolor}
\def\BibTeX{{\rm B\kern-.05em{\sc i\kern-.025em b}\kern-.08em
    T\kern-.1667em\lower.7ex\hbox{E}\kern-.125emX}}
\begin{document}

\title{Continuously Adapting Random Sampling (CARS) for Power Electronics Parameter Design\\
}

\author{\IEEEauthorblockN{
    Dominik Happel,
    Philipp Brendel, 
    Andreas Rosskopf,
    Stefan Ditze
}
\IEEEauthorblockA{
    Fraunhofer Institute for Integrated Systems and Device Technology IISB\\
    91058 Erlangen, Germany\\
    Email: \{dominik.happel, philipp.brendel, andreas.rosskopf, stefan.ditze\}@iisb.fraunhofer.de
}}
\maketitle

\begin{abstract}

To date, power electronics parameter design tasks are usually tackled using detailed optimization approaches with detailed simulations or using brute force grid search grid search with very fast simulations. A new method, named "Continuously Adapting Random Sampling" (CARS) is proposed, which provides a continuous method in between. This allows for very fast, and / or large amounts of simulations, but increasingly focuses on the most promising parameter ranges. Inspirations are drawn from multi-armed bandit research and lead to prioritized sampling of sub-domains in one high-dimensional parameter tensor. Performance has been evaluated on three exemplary power electronic use-cases, where resulting designs appear competitive to genetic algorithms, but additionally allow for highly parallelizable simulation, as well as continuous progression between explorative and exploitative settings. 
 
\end{abstract}



\section{Introduction}

Power electronics as research field is increasingly drawing the attention of applied Artificial Intelligence researchers. The Overview of Artificial Intelligence Applications in Power Electronics by Zhao et al \cite{zhao-2021}, points that the amount of research papers in this domain has been continuously increasing over the last years. Additionally, they provide three application categories corresponding to the lifecycle of power electronics systems: design process, control, and maintenance. 

The review by De León-Aldaco et al \cite{aldaco-2015} notes that “classical” optimization methods with optimality guarantees are often overwhelmed by the complexity of power electronic parameter spaces. Instead “metaheuristic” methods aim at "good enough" solutions, which enables them to effectively explore large search spaces, but which prevents arguments about solution completeness or optimality. Among metaheuristic methods, a separation can be performed into “trajectory-based” and “population-based” methods. Trajectory-based methods often perform local optimization, based on one solution, to quickly reach some local optima. Population based methods, which also include Genetic Algorithms (GAs) \cite{rosskopf-2018}, on the other hand consider multiple solutions simultaneously, often leading to better exploration.

A different argument is proposed by Guillod and Kolar \cite{guillod-2020}, who suggest that fast, vectorized simulations and brute force grid search should be preferred for magnetics design. Using a priori fixed grid sizes, this will eventually lead to many samples investigating non useful areas and only a few samples providing relevant results. This leaves engineers with a rough idea of interesting areas but requires another more fine-grained iteration to investigate subtle differences and progressions between relevant samples. 

A fast approach, considering all possible parameter settings, while increasingly focusing on the most relevant parameter ranges would be desirable. Similar problems are well known in reinforcement learning research, typically referred to as "exploration-exploitation dilemma" \cite{sutton}. Dividing parameter ranges in a fixed set of sub-domains, power electronics design tasks can be considered as multi-armed bandit problem, where the goal is to investigate those sub-domains providing the best results, while ensuring that no even better sub-domain is ignored. A typical solution to this problem is upper confidence bounds, which is often superior to other approaches when attempting to find one most promising sub-domain \cite{sutton}. However, since power electronics design is rather concerned with investigating most, if not all relevant sub-domains, a weighted softmax function \cite{softmax} can be considered instead. A similar stochastic sampling approach has been successfully applied by Schaul et. al. \cite{schaul} and additionally allows continuous progression between random sampling (exploration) and greedy focusing on a very few promising sub-domains (exploitation). Using fast, vectorized array operations \cite{numpy}, sampling speed is primarily limited by the number of sub-domains on which the softmax function is computed, which increases exponentially with the number of sampling parameters.

The sketched procedure presents similarities with Enhanced Continuous Tabu Search (ECTS) \cite{chelouah-2000}, where a search space is first divided into large global search areas, and later more focused search is performed in sub-domains of the most promising global search areas. Their comparison with other optimization approaches showed that their approach is quite competitive, especially for high dimensional parameter spaces (more than 10 parameters). However, this approach considers two steps of granularity, where either global diversification or local intensification steps are performed, which cannot be easily specified for arbitrary design tasks in advance. 

The proposed method, Continuously Adapting Random Sampling (CARS), is based on the described multi-armed bandit analogy, which provides less granularity than ECTS \cite{chelouah-2000} but allows arbitrary and especially continuous progression from exploration as random sampling, to exploitation of one single sub domain, just as ECTS intensification \cite{schaul}. Performance of CARS is evaluated on three exemplary power electronic use-cases and compared to GAs.

The paper is structured as follows, section \ref{sec:cars} describes the CARS method is detail. A brief overview of GAs is provided in section \ref{sec:ga}. A comparison of both methods, focused on sample evaluation is performed in section \ref{sec:methods}, as well as a CARS runtime study for increasing amounts of parameters and samples. The three power electronic design tasks are introduced, and comparison results are presented in section \ref{sec:use-case}. Conclusions are drawn in section \ref{sec:conclusion}, and finally limitations and possible directions for further research are discussed in the last section, \ref{sec:discussion}.


\section{Continuously Adapting Random Sampling} 
\label{sec:cars}

The basic idea of the proposed CARS method is to use a scalar preference metric for sub-domains of specified parameter ranges, to determine which sub-domains are "interesting" (should be investigated more often) and which are "uninteresting" (should be investigated less often). In reinforcement learning research this preference measure is referred to as "reward" \cite{sutton}, for GAs, the term "fitness" is established \cite{nsga2}. This paper uses the term "fitness", since GAs are considered for comparison. Fitness computation details for CARS and GAs are provided in section \ref{sec:methods}.

Initially, before any CARS sampling can be performed, a specified number of parameters \textit{$N_{param}$}, is divided into a number of sub-domains \textit{$N_{subdomain}$}, resulting in a tensor with \textit{$N_{param}$} dimensions of \textit{$N_{subdomain}$} elements. Inspired by optimistic initialization \cite{sutton}, a fitness value of \textit{$0.75$} is assigned to all sub-domains, which represents an above average fitness considering fitness normalization to approximately \textit{$[0, 1]$}, as described in section \ref{subsec:fitness}. 

The actual CARS procedure starts with computing a weighted softmax function \cite{softmax} \eqref{softmax} on the flattened sub-domain tensor \cite{numpy} of size \textit{$N_{tensor}$} \eqref{tensor-size}. This represents probabilities for sampling any sub-domain, identified by its index \textit{$j$}. The softmax weighting factor \textit{$\alpha$} is used for shifting the softmax focus from exploration to exploitation (or vice versa).

\begin{equation}
\label{tensor-size}
N_{tensor}=N_{subdomain}^{N_{param}}.
\end{equation}

\begin{equation}
\label{softmax}
P(\text{sub-domain}=j) = \frac{e^{fitness_j \cdot \alpha}}{\sum^{N_{tensor}}_{k=1} e^{fitness_k \cdot \alpha}}.
\end{equation}

Since sub-domains represent parameter ranges instead of specific parameter-values, samples are selected uniformly random within sub-domains. 

After simulation, the fitness value (computed according to \ref{subsec:fitness}) is associated with its initially selected tensor position \textit{$j$} and in case sub-domains have been investigated multiple times, the respective maximum is chosen.
The assigned fitness values directly affect sub-domain probabilities after softmax computation, where large fitness values are sampled more frequently, while initial fitness values (unexplored sub-domains) are still more probable than very poorly evaluated sub-domains. 

\begin{figure*}[htb!]
    \centering
    \includegraphics[width=1.0\linewidth]{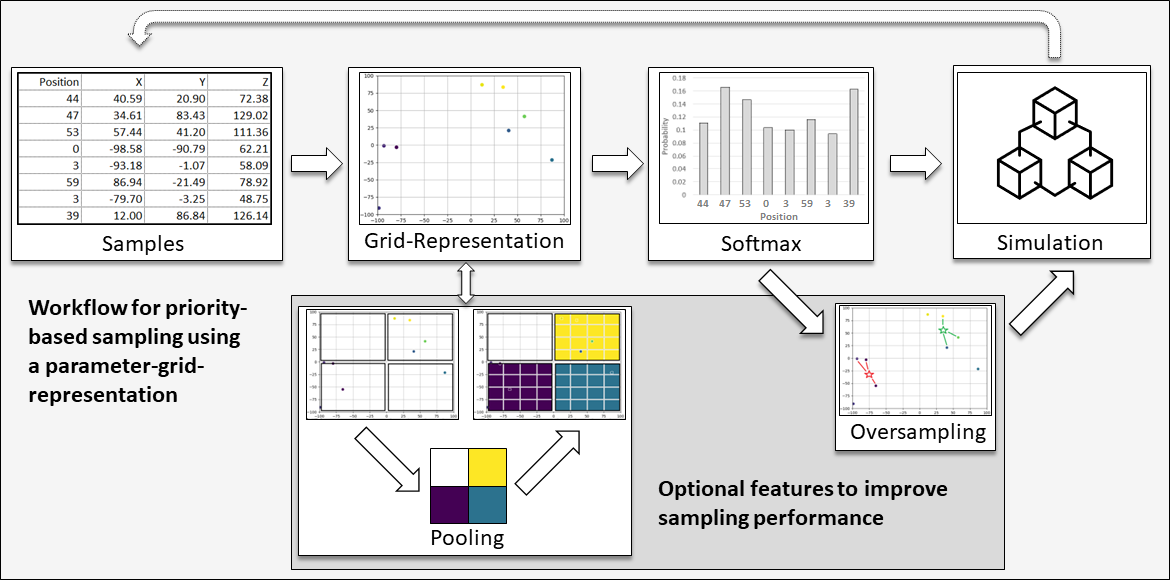}
    \caption[CARS Overview]
	{CARS method overview: Top row shows the main process, where fitness values computed in previous simulations are assigned to their sub-domain (according to their positional index \textit{$j$}). Here, variable \textit{$Z$} represents a non-normalized fitness measure. A weighted softmax function \eqref{softmax} is computed, to provide probabilities for sampling individual sub-domains for the next simulation iteration (qualitative results for exemplary sub-domains are shown). The bottom row shows extensions to distribute information to neighboring areas (pooling) and prevent unnecessary simulations based on their local fitness estimates (oversampling).}
    \label{figure:cars-overview}
\end{figure*}

The top row in figure \ref{figure:cars-overview} shows this procedure for an intermediate iteration, which starts with assigning previous simulation results to their tensor sub-domains according to positional index \textit{$j$}. Then, the softmax function \eqref{softmax} is computed for the whole tensor, converting fitness values to sub-domain sampling probabilities. Samples for the next simulation iteration are drawn uniformly random inside selected sub-domains and used for simulation.

Two additional extensions are implemented, to improve the information gain by distributing information to adjacent sub-domains, and to prevent unnecessary simulations due to the fact of random sampling inside sub-domains.


\subsection{Extension: Max-Pooling}
\label{sec:pool}

Caused by the exponential relation between number of parameters \textit{$N_{param}$}, number of sub-domains \textit{$N_{subdomain}$}, and total number of tensor elements \textit{$N_{tensor}$} \eqref{tensor-size}, even minor problems do not necessarily provide fitness information for each sub-domain. For example \textit{$6$} parameters with \textit{$10$} sub-domains each, results in a tensor of one million (\textit{$10^6$}) sub-domains.

 Therefore, a max pooling \cite{max-pool} inspired pooling option is included, to combine \textit{$N_{pool}$} adjacent sub-domains for each parameter to their maximum value, resulting in a compressed tensor of size:

\begin{equation}
N_{pooltensor}=\left( \frac{N_{subdomain}}{N_{pool}} \right)^{N_{param}}.
\end{equation}

This compressed information is added to the initial \textit{$N_{tensor}$} tensor, where each fitness value is summed with the max-pooled value to which it contributed. The intuition is to allow some simple interpolation of neighboring sub-domains (if they are combined in the same max-pooled area), while simultaneously maintaining differences between sub-domains. Additionally, the max-pooling operation requires exponentially less operations compared to averaging operations on the initial tensor, only \textit{$N_{pooltensor}$} maximization operations precisely. This is presented in the lower part of figure \ref{figure:cars-overview}, where initially \textit{$64$} \textit{$(8^2)$} sub-domains are combined to only \textit{$4$} \textit{$(2^2)$} max-pooled values and added to the initial tensor.


\subsection{Extension: Oversampling}
\label{sec:oversampling}

Besides spreading information to neighboring sub-domains, also sample-efficiency can be essential. Since not all simulations can be performed with several thousand simulations per second, a fast approximation of expected fitness values can reduce simulation requirements. Therefore, fitness values for prospective samples are estimated using the scikit-learn implementation of k-nearest neighbor regression \cite{scikit-2011}. Since nearest neighbor regression does not build a predictive model, it can estimate fitness values solely based on existing samples and their distances to samples under consideration. 

This extension is especially useful for detailed circuit simulations, where one circuit simulation consumes much more time than multiple nearest neighbor fitness estimations. In this case, the CARS method does not just sample an array of \textit{$N_{samples}$}, but instead a matrix of \textit{$N_{samples} \times N_{oversampling}$}. These samples are estimated using nearest neighbor regression, and samples for simulation are selected row-wise, where the sample with highest fitness estimate out of \textit{$N_{oversampling}$} samples is chosen. For example, when selecting two samples with oversampling of three, \textit{$2 \times 3$} sample candidates are sampled and estimated using nearest neighbor regression. Based on the fitness estimate, the most promising samples are used for simulation, in this case each of the two rows selects one out of three candidates.


\subsection{Heuristics}
\label{heuristics}

Since electronics design engineers are not necessarily familiar with algorithmic details, additional experience driven heuristics have been implemented to allow configuration of the CARS method based on abstract high-level-settings. 

To ensure fast sampling for potentially large tensors, and to reduce the "overhead per sample" for computing the softmax function on large tensors, CARS samples batches of \textit{$N_{samples}$} simultaneously. These details are not necessarily relevant for design engineers, therefore only the number of total samples \textit{$N_{total}$} is required, which is distributed in \textit{$N_{iteration}$} iterations of \textit{$N_{samples}$} samples each. To prevent extremely odd iteration numbers, they are preferably assigned as a multiple of \textit{$5$}.

\begin{equation}
\begin{aligned}
N_{iter} = floor(\frac{N_{total}^{\frac{2}{5}}}{5}) \cdot 5  \qquad & if N_{iter} \geq 5 \\
N_{iter} = floor(N_{total}^{\frac{2}{5}})  \qquad & otherwise\\.
\end{aligned}
\end{equation}

\begin{equation}
N_{samples} = floor(\frac{N_{total}}{N_{iter}}).
\end{equation}

Similarly, parameter ranges (\textit{$N_{param}$}) are by default divided into \textit{$9$} sub-domains (\textit{$N_{subdomain}$}) and pooling is performed by combining \textit{$3$} adjacent sub-domains of each parameter (\textit{$N_{pool}$}). 
By default, the softmax weighting \textit{$alpha$} increases by \textit{$1$} for each iteration, starting with initially random sampling (\textit{$\alpha = 0$}).

Oversampling, settings are also inferred based on the number of parameters \textit{$N_{param}$}, representing the dimensionality of the sampling space. If sufficient samples are available, the number of neighbors for nearest neighbor regression is computed as \eqref{neighbor}. Otherwise, all available samples are utilized.

\begin{equation}
\label{neighbor}
N_{neighbor}=(N_{param} \cdot 2)+1.
\end{equation}

The number of additional samples used for oversampling \textit{$N_{oversampling}$} is computed according to:

\begin{equation}
N_{oversampling}=N_{param}^{\frac{3}{2}}.
\end{equation}

This leaves engineers with a need to specify only the number of total samples \textit{$N_{total}$} as well as a set of parameters \textit{$N_{param}$} and objectives, while other settings can be inferred automatically.


\section{Genetic Algorithms}
\label{sec:ga}

GAs are frequently used to solve many engineering problems \cite{zhao-2021}, \cite{Fühner-2007}, \cite{Man-1996}, \cite{Johnson-1997} and are therefore used for benchmarking against the proposed CARS method.
GAs are based on the idea of evolving populations consisting of individuals via the three basic principles (i) mutation, (ii) crossover and (iii) selection.
In the context of this paper, each individual represents a set of circuit simulation parameters that are randomly mutated, or partially exchanged ("crossover") with another individual during evolution.
After a certain number of iterations, the algorithm selects certain individuals based on their fitness-values, representing the goals of optimization.
For this selection process, we use an adaption of the Non-dominated Sorting Genetic Algorithm "NSGA") \cite{nsga}, that was later referred to as NSGA-II by Deb et al. \cite{nsga2}.
NSGA's selection is guided by non-dominated sorting, i.e. an individual is preferably selected, if it is pareto-optimal (there is no other individual that is better in one objective and not worse in all other objectives) and a sharing-method promoting diversity in a population and preventing convergence to single pareto-optimal points, a known caveat of GAs in general, cf. e.g. \cite{schaffer}. For more details on the specific adaptions introduced by NSGA-II, we refer to the original work in \cite{nsga} and \cite{nsga2}, respectively. This concept is even more powerful, when running multiple parallel GAs on separate "islands" \cite{rosskopf-2018}. On the one hand, this allows efficient usage of High-Performance-Computing (HPC) resources, to perform simulations for multiple islands in parallel. On the other hand, it enables the exploration of different evolution strategies that are ultimately re-combined by selecting the best individuals among all islands.
More Details about the GA setup as well as the used mutation and crossover techniques are presented in appendix \ref{sec:ga_params}.


\section{Method overview}
\label{sec:methods}

\subsection{Fitness computation}
\label{subsec:fitness}

Both methods, CARS and GAs are designed to prioritize boundary conditions over objectives \cite{Jordehi-2015}. Therefore, a penalty term is computed for parameter settings that do not meet the boundary condition criteria according to the following formula, based on the square root of the Canberra distance:

\begin{equation}
\label{penalty}
 penalty = \rho * \sqrt{\frac{|value-target|}{|value|+|target|}}.
\end{equation}

\begin{figure}[htb!]
    \centering
    \includegraphics[width=1.0\linewidth]{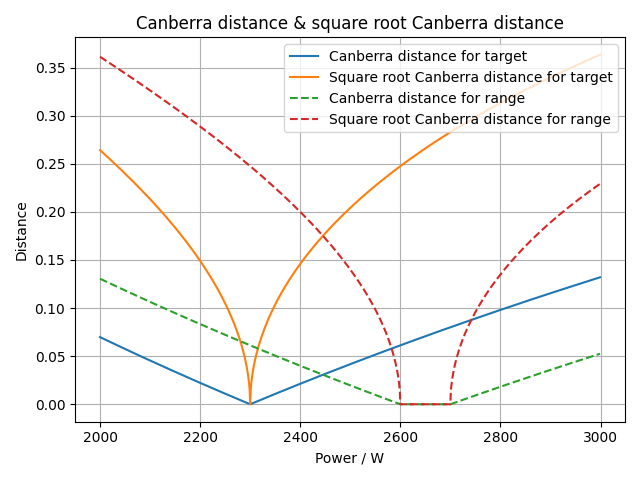}
    \caption
	{Square root Canberra Distance shows a larger increase of its distance measure, exemplary presented for a "target" value of 2300W (solid) and a "range" of 2600W to 2700W (dashed).}
    \label{figure:canberra-distance}
\end{figure}

Priorization is achieved by applying a weighting factor \textit{$\rho = 100$} for CARS and \textit{$\rho = 10,000$} for GAs. The square root of the Canberra distance is used, to ensure relatively large penalties (derived from the distance as in \eqref{penalty}) even for minor deviations. This is shown in figure \ref{figure:canberra-distance} for a target power value (solid lines) and for a power range (dashed lines).

Individual objective fitness values depend on the objective type, where the measurement value itself is used for "maximization" objectives. For "minimization" objectives, it is the negative value. The square root of the Canberra distance (solid line in Figure \ref{figure:canberra-distance}) with a weight of \textit{$\rho=-1$} is used for "target" objectives.
Finally, a "minimize range" objective is introduced, which considers potentially multiple operating points simultaneously. This fitness measure computes the difference of the largest and the smallest measurement values for \textit{$n$} operating points, according to \eqref{min-fsw}.

\begin{equation}
\label{min-fsw}
\begin{gathered}
range = - max(meas_1, ..., meas_n) \\ + min(meas_1, ..., meas_n).
\end{gathered}
\end{equation}

CARS considers a scalar fitness value for each sub-domain; therefore, the mean fitness is computed for all objectives, which is subtracted by the mean penalty for all boundary conditions. In contrast, GAs consider objectives individually and therefore the penalty term for each individual boundary condition is subtracted from all objective-fitness-values independently.

Since CARS is performed in batches, it offers the possibility to automatically scale fitness values for objectives or penalties for boundary conditions to approximately \textit{$[0, 1]$} based on values encountered during the first batch. Therefore, minimum, and maximum values of the first batch \textit{$vals_{first}$} are stored and used for normalizing current batches \textit{$vals_{newest}$} \eqref{normalize}. This normalization is computed for every boundary condition and objective independently, while all operating points are considered simultaneously. To prevent overflows caused by extremely large softmax exponents, the resulting scalar fitness value is normalized as well. However, the maximum fitness value is usually slightly larger than \textit{$1.0$}, since the overall best samples are seldom discovered during the first batch.

\begin{equation}
\label{normalize}
normalized = \frac{vals_{newest}-min(vals_{first})}{max(vals_{first})-min(vals_{first})}.
\end{equation}


\subsection{CARS runtime}
\label{sec:runtime}

Since CARS aims to improve on brute force grid search for very fast simulations \cite{guillod-2020}, the CARS runtime is investigated for batches of up to one million samples. This includes assigning previous simulations to their respective sub-domain, softmax probability computation, max-pooling, and fitness computation, for different amounts of parameters (\textit{$N_{param}$}) and batch sizes (\textit{$N_{samples}$}). The simulation time and the oversampling extension are excluded for this test, since they would scale linearly with the number of samples, and nearest neighbor estimation provides little benefit when performing very fast simulations. The simulation itself could also be distributed to multiple HPC machines, allowing highly parallelizable execution.

\begin{figure}[htb!]
    \centering
    \includegraphics[width=1.0\linewidth]{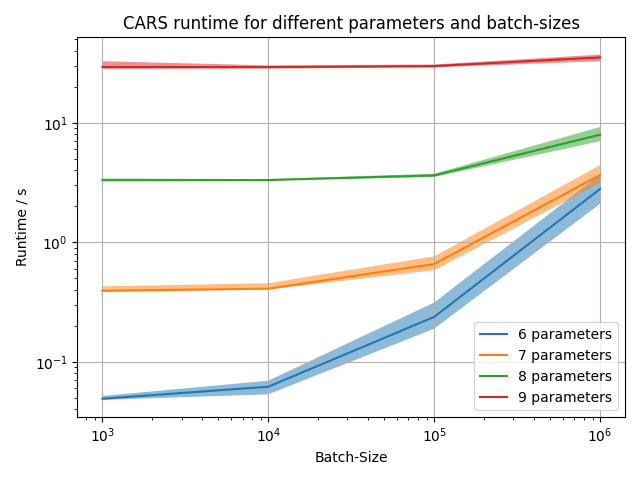}
    \caption
	{CARS runtime for different batch-sizes and numbers of sampling parameters, with each parameter divided into \textit{$9$} sub-domains. The CARS method is able to suggest up to one million samples in the time span of a few seconds, allowing focusing sampling in combination with very fast simulations.}
    \label{figure:runtime}
\end{figure}

For each test, \textit{$20$} consecutive sampling iterations, performed on one core of an Intel Xeon Gold 6134 Server CPU, are investigated and reported with mean (solid line), as well as min and max values (shaded areas) in figure \ref{figure:runtime}. As expected, especially for many parameters, larger batch-sizes show less distinct increases in runtime, since the essential softmax operation on the very large sub-domain tensors needs to be computed only once. For \textit{$9$} parameters, the resulting 9-dimensional tensor, with by default \textit{$9$} sub-domains each, contains more than \textit{$387$} million sub-domains (\textit{$9^9$}). These results suggest CARS can serve fast simulations with thousands or even one million samples in terms of a few seconds. However, especially for many parameters it might be beneficial to reduce the default number of sub-domains per parameter, where a reduction to just \textit{$8$} sub-domains would reduce the tensor size to about \textit{$134$} million (\textit{$8^9$}).

\section{Power Electronic Use-Cases}
\label{sec:use-case}

To determine the performance of the proposed CARS method in section \ref{sec:cars}, a parameter design evaluation for three exemplary power electronic use cases is performed, and compared to GAs. A first proof-of-concept application is a textbook boost converter \ref{sec:boost}, the second is a reproduction of a previous LLC Resonant Converter optimization using GAs \cite{rosskopf-2018} \ref{sec:llc-2018}, and the third is a recent research question for a LLC Resonant Converter with large power range \ref{sec:llc-2022}.
These use cases are modelled as LTspice netlists and simulations are performed using the SchoKI framework \cite{schoki} for parallel computation of LTspice \cite{ltspice} simulations.


\subsection{Boost Converter}
\label{sec:boost}

\begin{figure}[htb!]
    \centering
    \includegraphics[width=1.0\linewidth]{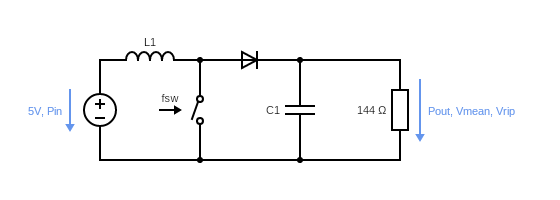}
    \caption
	{Schematic of Boost Converter.}
    \label{figure:boost-schematic}
\end{figure}

The circuit for the textbook Boost Converter is shown in Figure \ref{figure:boost-schematic}. Parameters, objectives and boundary conditions are listed in Table \ref{table:param-boost}. This application includes only one operating point and a total of \textit{$5,000$} samples are investigated for each method. \textit{$log$} indicates sampling in logarithmic scale. Further details on measurements are provided in the appendix \ref{circuit-details}. Note that this first proof of concept example does not include any component loss models.

\begin{table}[htb!]
    \centering
    \begin{tabular}{ | c | c | c | c | }\hline
    
    \multicolumn{4}{|c|}{\textbf{Parameters}} \\\hline
    \textbf{Name}   & \textbf{Type}   & \textbf{Values}    & \textbf{Operating Points}  \\\hline
    C1           & log             &  [1e-9, 1e-3]      & 1  \\\hline
    L1           & log             &  [1e-6, 100e-3]    & 1  \\\hline
    fsw             & log             &  [100, 1e6]        & 1  \\\hline\hline

    \multicolumn{4}{|c|}{\textbf{Objectives}} \\\hline
    \textbf{Name}   & \textbf{Type}   & \textbf{Values}    & \textbf{Operating Points}  \\\hline
    vmean           & target          & 12                 & 1 \\\hline
    eff\_tot         & max             &   -                & 1\\\hline\hline

    \multicolumn{4}{|c|}{\textbf{Boundary Conditions}} \\\hline
    \textbf{Name}   & \textbf{Type}   & \textbf{Values}    & \textbf{Operating Points}  \\\hline
    vmean           & range           & [11.5, 12.5]       & 1\\\hline
    vrip            & range           &  [0, 2]            & 1 \\\hline
    
    \end{tabular}
    \smallskip
    \caption
    {Parameters, Objectives, and Boundary Conditions for Boost Converter.}
    \label{table:param-boost}
\end{table}

\begin{figure}[htb!]
    \centering
    \includegraphics[width=1.0\linewidth]{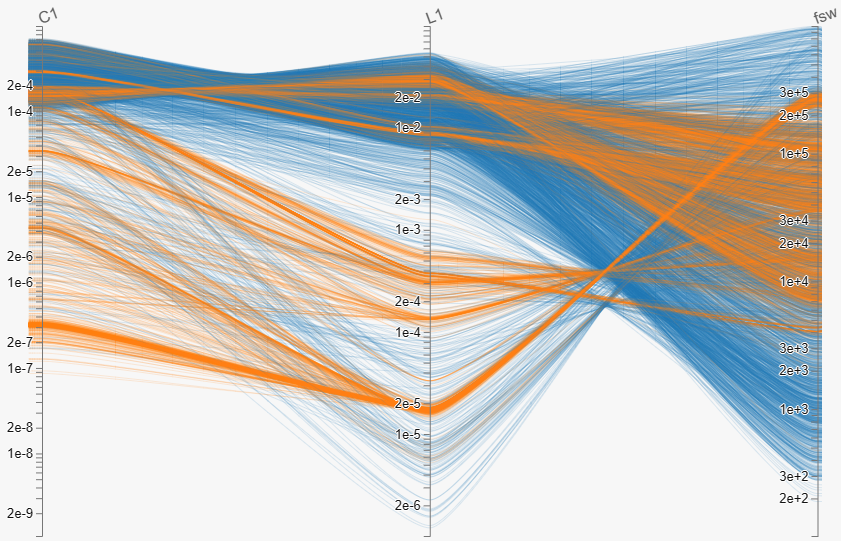}
    \caption
	{Parameter values of samples that satisfied Boost Converter boundary conditions (CARS blue, GA orange).}
    \label{figure:param-boost}
\end{figure}

\begin{figure}[htb!]
    \centering
    \includegraphics[width=1.0\linewidth]{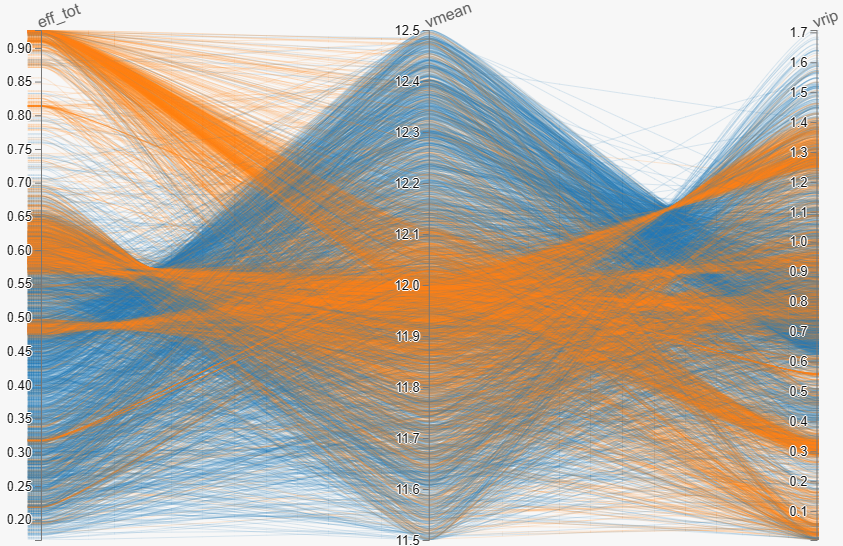}
    \caption
	{Simulation results for samples that satisfied Boost Converter boundary conditions (CARS blue, GA orange)}
    \label{figure:obj-boost}
\end{figure}

The CARS method found in total \textit{$2,477$} samples satisfying all boundary conditions, while GAs found \textit{$3,367$}. Valid parameters in figure \ref{figure:param-boost} show for CARS primarily one large, densely sampled parameter range. The resulting, valid GA samples cover a wider parameter range, but samples tend to be grouped in distinct clusters. 
Regarding measurements, presented in Figure \ref{figure:obj-boost}, highest "efficiencies" were achieved primarily by GA samples, the desired output voltage of \textit{$12V$} was reached by both methods, and a CARS samples achieved the lowest voltage ripple / overshoot. 

\subsection{LLC-2018}
\label{sec:llc-2018}

\begin{figure}[htb!]
    \centering
    \includegraphics[width=1.0\linewidth]{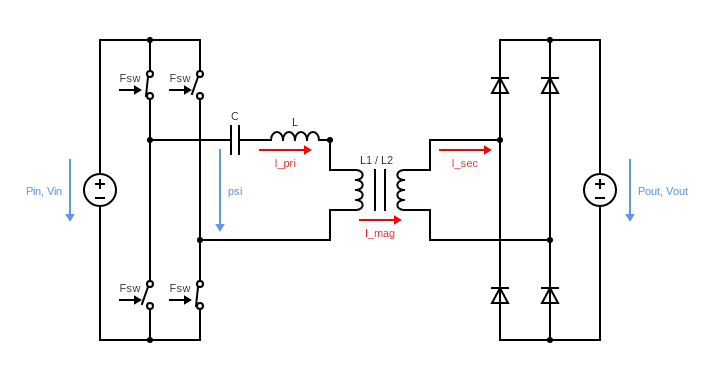}
    \caption
	{Exemplary schematic of an LLC resonant converter, detailed schematics including static loss models are provided in the appendix figures \ref{schematic-2018} and \ref{schematic-2022}.}
    \label{figure:llc-schematic}
\end{figure}

Figure \ref{figure:llc-schematic} shows a simplified LLC Resonant Converter, highlighting the most relevant components. The complete circuit for simulation is provided in the appendix, figure \ref{schematic-2018}. Parameters are listed in Table \ref{table:param-llc-2018}. This application considers five operating points, where hardware components are fixed, but five different switching frequencies are applied to reach different power levels for different output voltages. The output voltage is defined as categorical \textit{$grid$} parameter with only one possible value for each operating point. A total of \textit{$100,000$} samples are investigated for each method. Further details on measurements are provided in the appendix \ref{circuit-details}.

\begin{table}[htb!]
    \centering
    \begin{tabular}{ | c | c | c | c | }\hline

    \multicolumn{4}{|c|}{\textbf{Parameters}} \\\hline
    \textbf{Name}   & \textbf{Type}  & \textbf{Values}   & \textbf{Operating Points}  \\\hline
    C\_res          & log            & [1e-9, 1.6e-6]    & 1                 \\\hline
    L\_ser          & linear         & [5e-6, 40e-6]     & 1                 \\\hline
    L1              & log            & [50e-6, 1e-3]     & 1                 \\\hline
    n               & linear         & [0.6, 2.2]        & 1                 \\\hline
    f\_sw           & linear         & [50e3, 320e3]     & 5                 \\\hline
    \multirow{5}{*}{V\_out}  & \multirow{5}{*}{grid}     &  300  & \multirow{5}{*}{5} \\
                                &                           & 350   & \\
                                &                           & 400   & \\
                                &                           & 450   & \\
                                &                           & 500   & \\\hline\hline

    \multicolumn{4}{|c|}{\textbf{Objectives}} \\\hline
    \textbf{Name} & \textbf{Type} & \textbf{Values} & \textbf{Operating Points}  \\\hline
    i\_pri\_res\_rms      & min  & -  & 5 \\\hline
    i\_sec\_res\_rms      & min  &   -  & 5 \\\hline
    \multirow{5}{*}{p\_out\_avg}               & \multirow{5}{*}{target}  &  2660 & \multirow{5}{*}{5} \\
    & & 3,160 & \\
    & & 3,660 & \\
    & & 3,660 & \\
    & & 3,660 & \\\hline\hline

    \multicolumn{4}{|c|}{\textbf{Boundary Conditions}} \\\hline
    \textbf{Name} & \textbf{Type} & \textbf{Values} & \textbf{Operating Points}  \\\hline
    psi\_grad       & range    & [10, 90]  & 5\\\hline
    \multirow{5}{*}{p\_out\_avg}      & \multirow{5}{*}{range}  &  [1995, 3325] & \multirow{5}{*}{5} \\
    & & [2,370, 3,950] & \\
    & & [2,745, 4,575] & \\
    & & [2,745, 4,575] & \\
    & & [2,745, 4,575] & \\\hline
    
    \end{tabular}
    \smallskip
    \caption
    {Parameters, Objectives, and Boundary Conditions for LLC-2018.}
    \label{table:param-llc-2018}
\end{table}

\begin{figure}[htb!]
    \centering
    \includegraphics[width=1.0\linewidth]{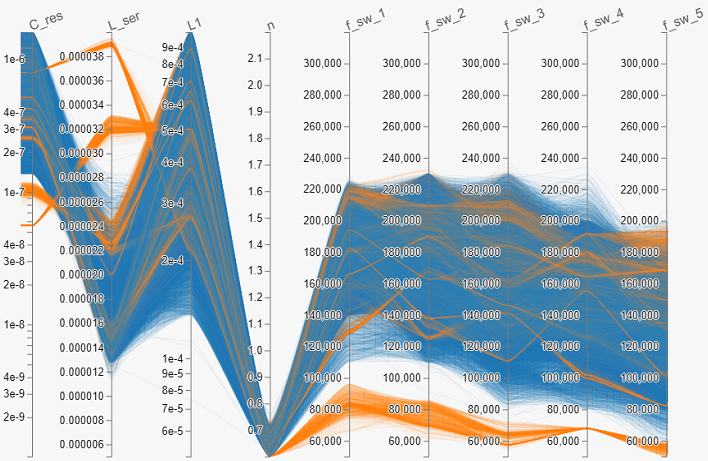}
    \caption
	{Parameter values of samples that satisfied LLC-2018 boundary conditions (CARS blue, GA orange)}
    \label{figure:param-2018}
\end{figure}

\begin{figure}[htb!]
    \centering
    \includegraphics[width=1.0\linewidth]{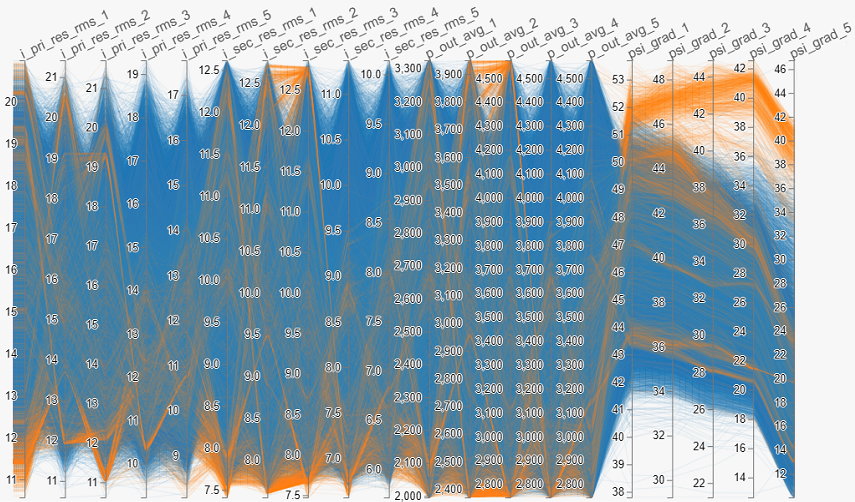}
    \caption
	{Simulation results for samples that satisfied LLC-2018 boundary conditions (CARS blue, GA orange)}
    \label{figure:obj-2018}
\end{figure}

In contrast to the previous test, for LLC-2018, CARS achieved \textit{$48,501$} samples satisfying all boundary conditions, and GAs \textit{$11,684$}. Regarding valid parameter ranges, shown in figure \ref{figure:param-2018}, CARS samples again tend to be grouped into one large parameter range, while GA samples form multiple sample clusters. Measurements, presented in figure \ref{figure:obj-2018}, do not allow for a simple comparison, since overall performance would require an averaged measure considering all operating points. But the results indicate that "locally" best performance for one of the operating points was partially achieved by CARS samples and partially by GA samples. It is however worth noting, that GAs discovered additional clusters for large \textit{$L\_ser$} and small \textit{$f\_sw$} values that were not considered by CARS. 

\subsection{LLC-2022}
\label{sec:llc-2022}

The third use-case, LLC-2022, shown in detail in Figure \ref{schematic-2022} in the appendix, considers three operating points simultaneously, and a large output power range, as listed in Table \ref{table:param-llc-2022}. Again, \textit{$100,000$} samples are evaluated for each method. The third objective \textit{$fsw$} accounts for the fact that large power ranges shall be covered and encourages small switching frequency deviations, which is computed according to \eqref{min-fsw} for all operating points simultaneously.

\begin{table}[htb!]
    \centering
    \begin{tabular}{ | c | c | c | c |}\hline

    \multicolumn{4}{|c|}{\textbf{Parameters}} \\\hline
    \textbf{Name}   & \textbf{Type}  & \textbf{Values}        & \textbf{Operating Points}  \\\hline
    L\_st1       & log            & [2e-6, 200e-6]         & 1  \\\hline
    A\_L           & log            & [20e-9, 500e-9]        & 1  \\\hline
    np              & log            &  [2, 200]              & 1  \\\hline
    ns              & log            &  [2, 200]              & 1  \\\hline
    C\_res1      & log            &  [1e-9, 100e-9]        & 1  \\\hline
    fsw             & linear         &  [100e3, 1e6]          & 3 \\\hline
    \multirow{3}{*}{V\_output}    & \multirow{3}{*}{grid}  &  850 & \multirow{3}{*}{3} \\
    & & 840 & \\
    & & 600 & \\\hline\hline

    \multicolumn{4}{|c|}{\textbf{Objectives}} \\\hline
    \textbf{Name} & \textbf{Type} & \textbf{Values} & \textbf{Operating Points}  \\\hline
    eta        & max    & -   & 3\\\hline
    mag\_current\_rms      & min  &   -  & 3 \\\hline
    fsw      & $\text{min}_{\text{range}}$  &  -  & all \\\hline\hline

    \multicolumn{4}{|c|}{\textbf{Boundary Conditions}} \\\hline
    \textbf{Name} & \textbf{Type} & \textbf{Values} & \textbf{Operating Points}  \\\hline
    psi\_deg        & range    & [5, 85] & 3 \\\hline
    eta      & range  &  [0.75, 1.0]  & 3 \\\hline
    \multirow{3}{*}{output\_power\_avg}      & \multirow{3}{*}{range}  &  [100, 300] & \multirow{3}{*}{3} \\
    & & [3,600, 4,000] & \\
    & & [3,000, 3,400] & \\\hline
    i\_swprimoff      & larger  &  0 & 3 \\\hline

    \end{tabular}
    \smallskip
    \caption
    {Parameters, Objectives, and Boundary Conditions for LLC-2022.}
    \label{table:param-llc-2022}
\end{table}

\begin{figure}[htb!]
    \centering
    \includegraphics[width=1.0\linewidth]{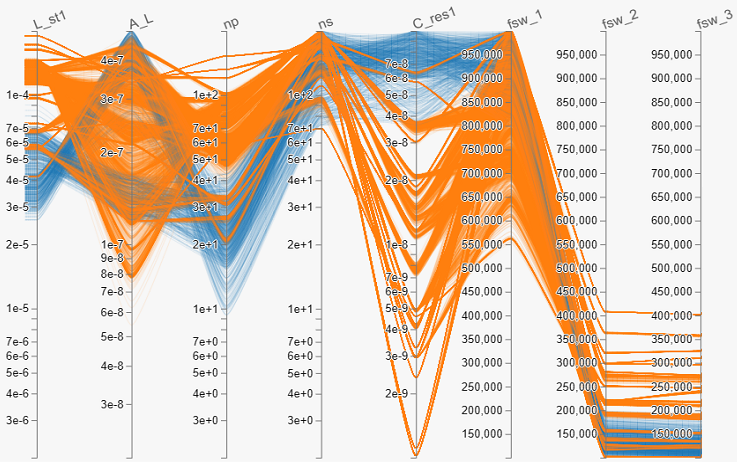}
    \caption
	{Parameter values of samples that satisfied LLC-2022 boundary conditions (CARS blue, GA orange)}
    \label{figure:param-2022}
\end{figure}

\begin{figure}[htb!]
    \centering
    \includegraphics[width=1.0\linewidth]{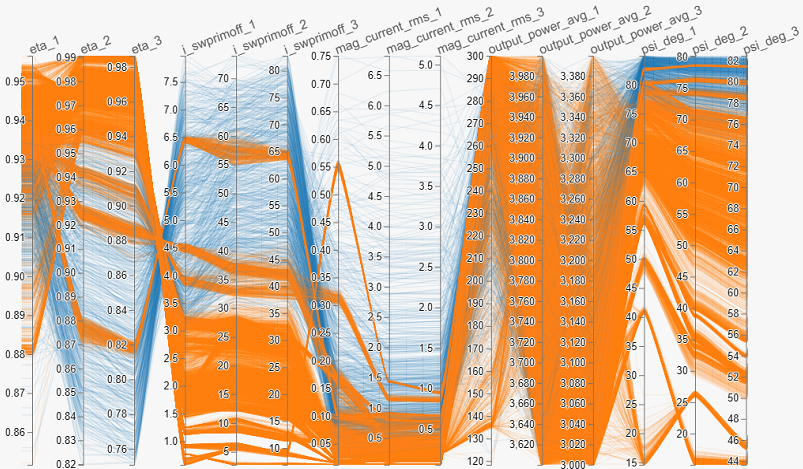}
    \caption
	{Simulation results for samples that satisfied LLC-2022 boundary conditions (CARS blue, GA orange)}
    \label{figure:obj-2022}
\end{figure}

This third use case presents \textit{$719$} samples satisfying all boundary conditions for CARS, in comparison to \textit{$14,358$} for GAs. Valid parameters, presented in figure \ref{figure:param-2022} show two smaller parameter ranges of valid samples for CARS and multiple separate clusters for GAs. Measurements, presented in figure \ref{figure:obj-2022} suggest superior performance for GAs, with larger efficiency values \textit{$eta$} usually achieved by GAs, as well as lower transformer currents \textit{$mag\_current\_rms$}. 

\subsection{CARS extension study}
\label{sec:extension_study}

\begin{figure}[htb!]
    \centering
    \includegraphics[width=1.0\linewidth]{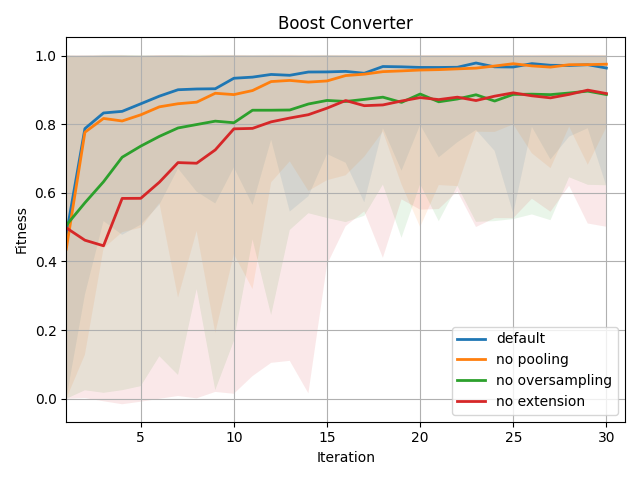}
    \caption[extension-boost]
	{Effect of extensions (pooling and oversampling) on sampled fitness for Boost Converter.}
    \label{extension-boost}
\end{figure}

\begin{figure}[htb!]
    \centering
    \includegraphics[width=1.0\linewidth]{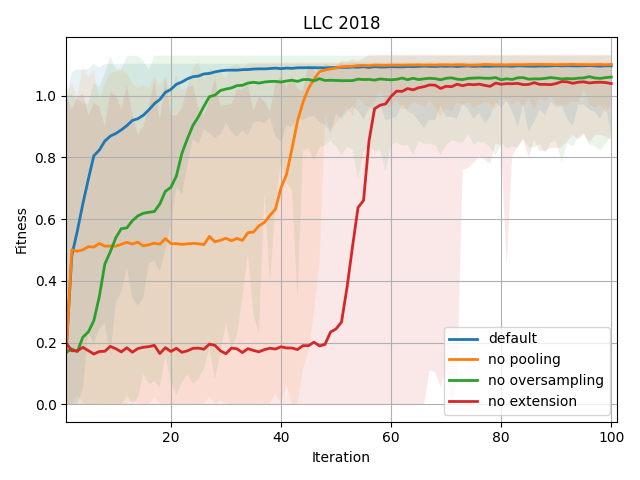}
    \caption[extension-2018]
	{Effect of extensions (pooling and oversampling) on sampled fitness for LLC-2018.}
    \label{extension-2018}
\end{figure}

To investigate performance benefits due to the extensions described in sections \ref{sec:pool} and \ref{sec:oversampling}, all three use cases were investigated without extensions for comparison. The minimum, maximum and mean fitness value for the default setting described in section \ref{heuristics} with both extensions, using oversampling but no pooling, using pooling but no oversampling and using no extension are presented in figures \ref{extension-boost}, \ref{extension-2018} and \ref{extension-2018} respectively.
Note that fitness values were normalized based on the maximum and minimum values sampled during the first iteration and may therefore exceed the previously best samples with fitness \textit{$1.0$}. Similarly, fitness values itself do not allow for performance comparisons since normalization constants vary between different CARS iterations.

Especially for the more complex use cases LLC-2018 and LLC-2022 the advantages of the extensions are well observable. As intended, pooling leads to faster increase in sampled fitness values since approximated fitness information is distributed to adjacent sub-domains. Similarly, oversampling leads to less variance, since randomly selected samples with poor neighbor-based fitness estimates are omitted from simulation. 
Finally, even tests using no extension suggest that softmax based sampling with increasing $\alpha$ prioritization can suggest useful samples. Assuming representative samples during the first batch (which is performed as random sampling, due to constant sub-domain fitness initialization), the increase in mean fitness indicates more emphasis on relevant parameter areas compared to random sampling.

\begin{figure}[htb!]
    \centering
    \includegraphics[width=1.0\linewidth]{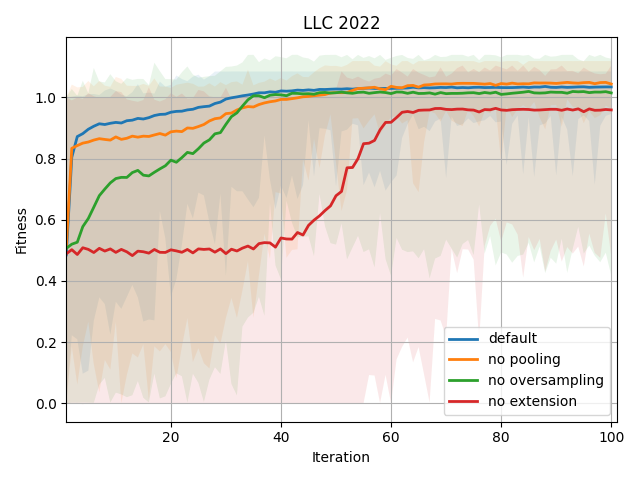}
    \caption[extension-2022]
	{Effect of extensions (pooling and oversampling) on sampled fitness for LLC-2022.}
    \label{extension-2022}
\end{figure}


\section{Conclusion}
\label{sec:conclusion}

Overall, the proposed CARS method is able to suggest valid parameter settings, satisfying all boundary conditions for all three power electronic use cases. Arguments regarding superiority of any method seem not justified based on the exemplary applications. GAs might reach slightly better objective values, while CARS suggests more diverse and continuous ranges of valid samples. However, there are some recurring properties of both methods: CARS tends to thoroughly investigate large parameter ranges, likely related to highly rated and therefore frequently sampled sub-domains of the parameter tensor. In contrast, GAs tend to find valid samples in small clusters, which is likely related to multiple parallel GAs sampling independently on distinct "islands". GAs occasionally provides samples in parameter ranges that were not successfully investigated by CARS, but CARS indicates that valid samples exist also between clusters found by GAs. This highlights the fact that in contrast to globally optimal approaches, neither of both methods can find all valid samples. 

An additional advantage of the proposed CARS method is fast sampling of large sample batches (up to a million samples) in a few seconds, which is advantageous for very fast simulations or in cases where time consuming simulations can be performed on multiple HPC machines in parallel. Additional extensions allow CARS to investigate large parameter ranges faster (pooling) and more focused (oversampling), while even softmax sampling itself seems to improve on random sampling. Finally, sub-domain indices of CARS samples can be stored alongside simulated samples, to re-establish previous parameter tensors and to perform additional simulations using for example specialized explorative or exploitative softmax weighting settings.


\section{Discussion}
\label{sec:discussion}

As previous conclusions highlight different advantages and drawbacks for CARS and GAs, further research might investigate a combination of both methods. Well performing parameter settings discovered by GAs could be associated with their respective CARS parameter tensor sub-domain. In this way the ability of GAs to find many locally clustered solutions can be combined with the ability of CARS to investigate large areas around well-performing samples thoroughly.

The suggested square-root Canberra distance measure has nice properties for the considered use-cases, but it will saturate at a distance of \textit{$1$} for negative values, which leads to no information regarding closeness. Applications including negative target values should consider this fact and use another distance metric.

To ease the application for electronic engineers, both optimization strategies, CARS and GAs used experience driven heuristics for various hyperparameters. This led to reasonable results for the considered use cases, but it is also likely that application specific hyperparameter settings would yield different and eventually better results. The extension study \ref{sec:extension_study} highlights an interesting situation, where especially for the more complex use-cases (LLC-2018 and LLC-2022) there are periods of larger "fitness slope". Since the prioritization parameter \textit{$\alpha$} corresponds by default to the iteration, this increase in average fitness seems related to a combination of the prioritization parameter \textit{$\alpha$} and the information density available in the parameter tensor, which is additionally increased by pooling and oversampling.

The three power electronic use-cases described in Section \ref{sec:use-case} were chosen to represent a variety of different parameter design applications, but it cannot be guaranteed that either of the investigated methods performs well on other applications. Additionally, the considered netlists consider simplified, static loss models or no loss models at all. Therefore, parameters suggested by any of the methods are unlikely "ideal" parameters for any of the use cases, but instead rather some initial suggestion for more detailed investigations.

Finally, the CARS method used a constant fitness value of \textit{$0.75$} as "optimistic initialization" of the parameter sub-domain tensor. It is therefore required to investigate all sub-domains individually (or approximate them by max pooling or oversampling). However, the parameter sub-domain tensor could also be initialized based on engineering knowledge about desirable and undesirable parameter ranges, which reduces the need for evaluating irrelevant areas through circuit simulation and might improve sample efficiency additionally.


\section*{Acknowledgment}

This work was supported by the Federal Ministry of Education and Research in the CODAPE "Kollaborative Entwicklungsumgebung für die Leistungselektronik" project via grant 16ME0356.



\clearpage
\section*{Appendix}


\subsection{GA settings}
\label{sec:ga_params}

Important parameters for genetic algorithms are the chosen number of islands, their population sizes, as well as the number of generations across which the populations are evolved. Table \ref{table:ga-settings-use-cases} lists the corresponding choices for the three considered use cases in this paper.

\begin{table}[htb!]
    \centering
    \begin{tabular}{ | c | c | c | c | }\hline

    \multicolumn{4}{|c|}{\textbf{GA Settings}} \\\hline
    \textbf{Use-Case}   & \textbf{\# Islands}  & \textbf{Population size}   & \textbf{Generations}  \\\hline
    Boost Converter & 5           & 20    & 50                 \\\hline
    LLC-2018        & 50          & 20    & 100                 \\\hline
    LLC-2022        & 50          & 20    & 100                 \\\hline 
    
    \end{tabular}
    \smallskip
    \caption
    {GA setups for the three power electronic use cases.}
    \label{table:ga-settings-use-cases}
\end{table}

Furthermore, the choice and probabilities for mutations and crossovers are decisive for a good trade-off between exploration of new areas in the parameter space and exploitation in the close surroundings of already evaluated individuals.
To allow partial evolution, mutation happens based on two probabilities \textit{$P_{mutate}$}, which determines whether an individual is mutated at all and \textit{$P_{mutate\_val}$}, which determines the likelihood of each individual value being mutated in case the individual is mutated at all. Mutation is based on a Gaussian distribution with standard deviation \textit{$\sigma_{mutate}$}.
Furthermore, we use simulated binary crossover with hyperparamaters \textit{$\alpha_{crossover}$} and \textit{$\eta_{crossover}$} determining the general resemblance of the product with respect to the two individuals used for crossover. We refer to \cite{simulated-binary} for more information on this crossover technique.
Table \ref{table:ga-settings} lists the parameters for mutation and crossover, which are promoting diversity within the populations and have proven superior for the considered use cases over less explorative configurations.

\begin{table}[htb!]
    \centering
    \begin{tabular}{ | c | c | }\hline

    \textbf{Parameter}   & \textbf{Value}   \\\hline
    $P_{\text{mutate}}$   &  0.1                      \\\hline
    $P_{\text{mutate\_val}}$   &  0.3             \\\hline
    $\sigma_{\text{mutate}}$   &  0.4         \\\hline 
    $P_{\text{crossover}}$  & 0.95  \\\hline
    $\alpha_{\text{crossover}}$ & 0.2     \\\hline
    $\eta_{\text{crossover}}$   & 1      \\\hline
    \end{tabular}
    \smallskip
    \caption
    {Hyperparameters used for Mutation and Crossover}
    \label{table:ga-settings}
\end{table}


\subsection{Circuit details}
\label{circuit-details}

For the textbook Boost Converter, presented in Section \ref{sec:boost}, the mean output voltage \textit{$vmean$} is obtained as mean voltage on the output load resistor after a startup phase of \textit{$10ms$}. The ripple / overshoot measure \textit{$vrip$} subtracts the mean output voltage \textit{$vmean$} from the maximum output voltage, also measured after a startup phase of \textit{$10ms$}. The simplified efficiency measure \textit{$eff\_tot$} divides total output power by the total input power. Since the schematic does not include component loss models, this efficiency measure primarily considers stored energy, e.g., in the output capacitor. This way there are competing objectives, where the \textit{$vrip$} measure requires large capacitors for smoothing, while the efficiency measure \textit{$eff\_tot$} suggests smaller capacitors to achieve large output powers.

\begin{figure*}[htb!]
    \centering
    \includegraphics[width=1.0\linewidth]{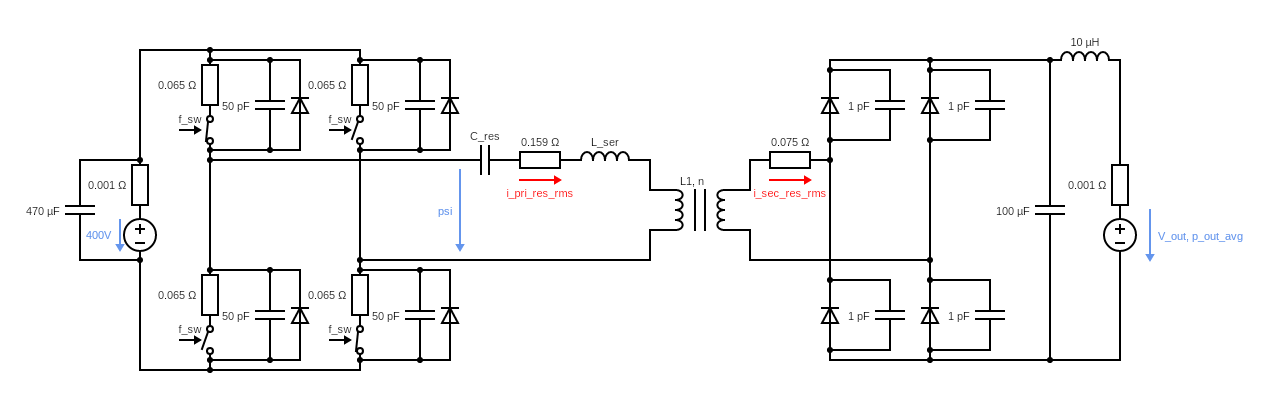}
    \caption[schematic-2018]
	{Schematic used for LLC-2018 simulations.}
    \label{schematic-2018}
\end{figure*}

Figure \ref{schematic-2018} provides the schematic used for LLC-2018 simulations. The measurements \textit{$i\_pri\_res\_rms$} and \textit{$i\_sec\_res\_rms$} are RMS current measurements on the respective components. Similarly output power \textit{$p\_out\_avg$} is the current and voltage product in the output sink. The phase shift measure \textit{$psi\_grad$} is used to determine whether zero voltage switching is possible. Therefore, the primary side zero-crossing is measured as \textit{$psi$} and \textit{$psi\_grad$} computed as:

\begin{equation}
psi\_grad=360 \cdot f\_sw \cdot psi.
\end{equation}

\begin{figure*}[htb!]
    \centering
    \includegraphics[width=1.0\linewidth]{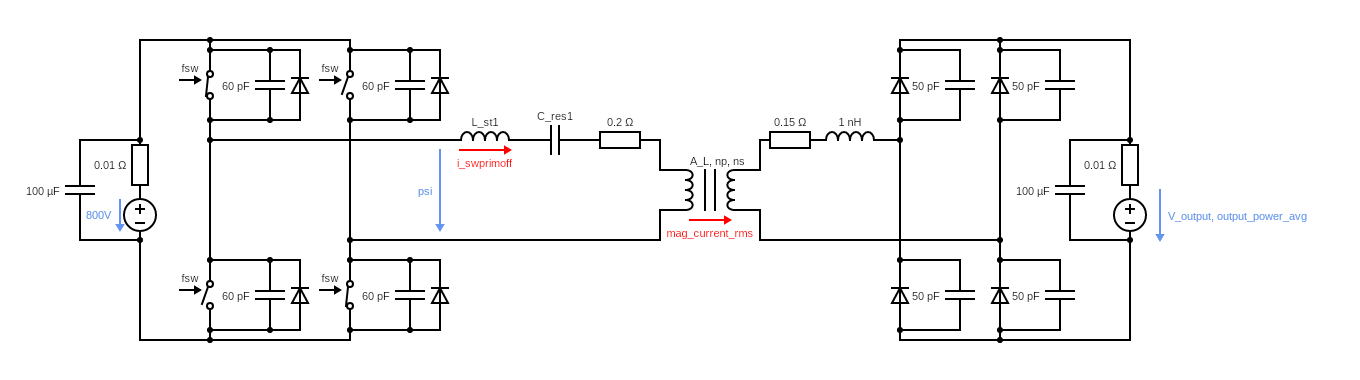}
    \caption[schematic-2022]
	{Schematic used for LLC-2022 simulations.}
    \label{schematic-2022}
\end{figure*}

Finally, figure \ref{schematic-2022} provides the schematic used for LLC-2022 simulations. The output power \textit{$output\_power\_avg$} is again computed as product of current and voltage in the output sink. The efficiency measure \textit{$eta$} is computed based on the output power \textit{$output\_power\_avg$} and static component losses. 

\begin{equation}
eta = \frac{output\_power\_avg}{output\_power\_avg+losses}.
\end{equation}

 The current \textit{$i\_swprimoff$} is measured in the respective component. The phase shift measure \textit{$psi\_deg$} is obtained by measuring the zero crossing of the primary voltage as \textit{$psi$} and computed as: 

\begin{equation}
psi\_deg = 360 \cdot fsw \cdot psi.
\end{equation}

The magnetic current \textit{$mag\_current\_rms$} is measured as current in the primary stray inductance subtracted by the turns-ratio factored current of the secondary stray inductance.

\begin{equation}
mag\_current\_rms = I(Lst1)-\frac{ns}{np}\cdot I(Lst2).
\end{equation}

\end{document}